\documentclass{bmvc2k}

\title{Centre Stage: Centricity-based\\Audio-Visual Temporal Action Detection}

\addauthor{Hanyuan Wang}{hanyuan.wang@bristol.ac.uk}{1}
\addauthor{Majid Mirmehdi}{majid@cs.bris.ac.uk}{1}
\addauthor{Dima Damen}{dima.damen@bristol.ac.uk}{1}
\addauthor{Toby Perrett}{toby.perrett@bristol.ac.uk}{1}

\addinstitution{
 Department of Computer Science\\
 Faculty of Engineering\\
 University of Bristol\\
 Bristol, UK
}

\runninghead{Wang et al}{Centricity-based Audio-Visual Temporal Action Detection}

\usepackage{graphicx}
\usepackage{amsmath}
\usepackage{amssymb}
\usepackage{booktabs}
\usepackage{times}
\usepackage{epsfig}
\usepackage{pifont}       
\usepackage{bbding}
\usepackage{fontawesome}
\usepackage{multirow}
\usepackage{amsfonts}
\usepackage{makecell,diagbox,verbatim,bm}
\usepackage{float}

\begin{document}

\maketitle

\begin{abstract}
Previous one-stage action detection approaches have modelled temporal dependencies using only the visual modality. In this paper, we explore different strategies to incorporate the audio modality, using multi-scale cross-attention to fuse the two modalities.
We also demonstrate the correlation between the distance from the timestep to the action centre and the accuracy of the predicted boundaries. Thus, we propose a novel network head to estimate the closeness of timesteps to the action centre, which we call the centricity score. This leads to increased confidence for proposals that exhibit more precise boundaries. Our method can be integrated with other one-stage anchor-free architectures and we demonstrate this on three recent baselines on the EPIC-Kitchens-100 action detection benchmark where we achieve state-of-the-art performance. Detailed ablation studies showcase the benefits of fusing audio and our proposed centricity scores. Code and models for our proposed method are publicly available at \url{https://github.com/hanielwang/Audio-Visual-TAD.git}.
\end{abstract}

\section{Introduction}
\label{sec:intro}
Temporal action detection aims to predict the boundaries of action segments from a long untrimmed video and classify the actions, as a fundamental step towards video understanding \cite{slowfast, I3D, TSN}. A typical challenging scenario is in unscripted actions in egocentric videos \cite{EPIC100, Ego4d} 
which contain dense action segments of various lengths in an unedited video, ranging from seconds to minutes.

\begin{figure}
\begin{tabular}{cc}
\bmvaHangBox{\includegraphics[width=0.49\linewidth]{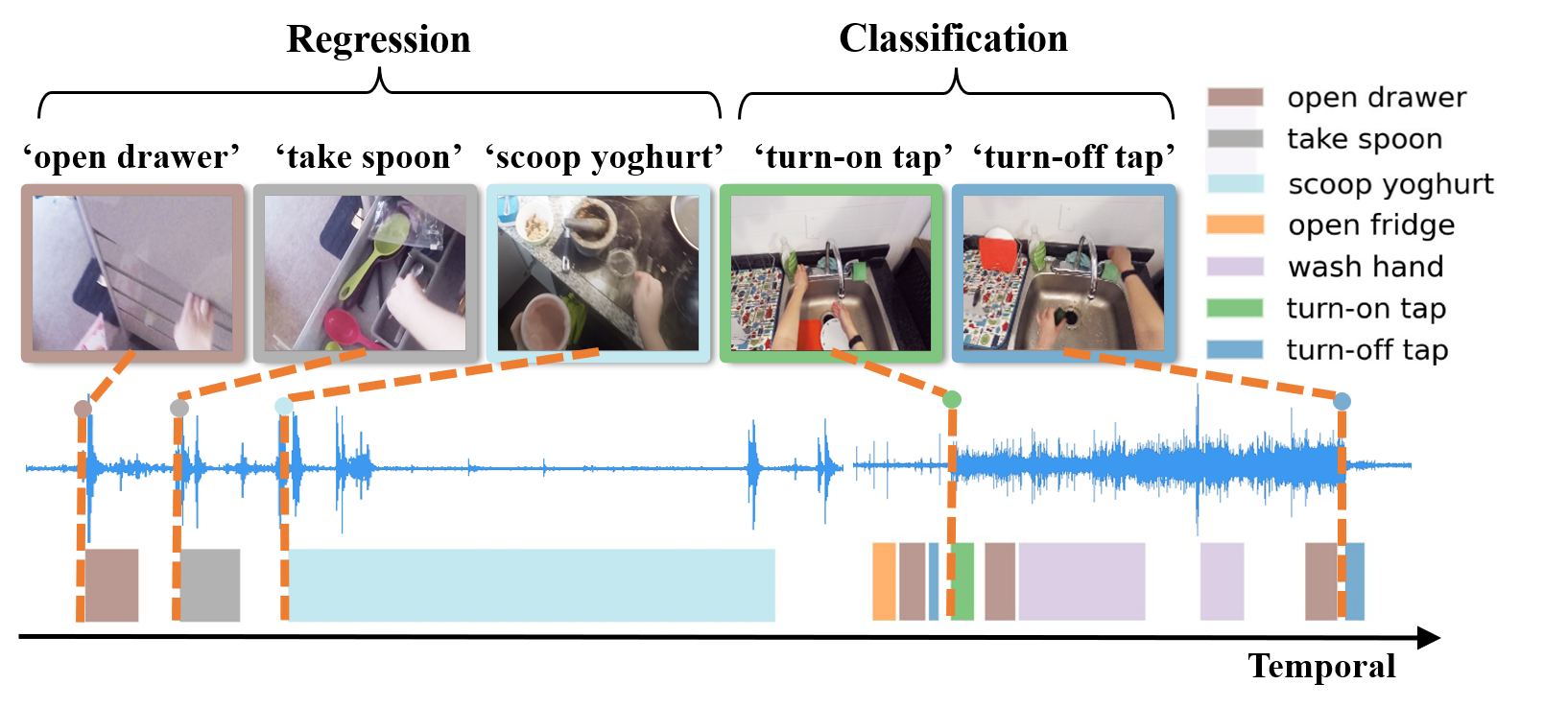}}&\hspace{-5mm}
\bmvaHangBox{\includegraphics[width=0.49\linewidth]{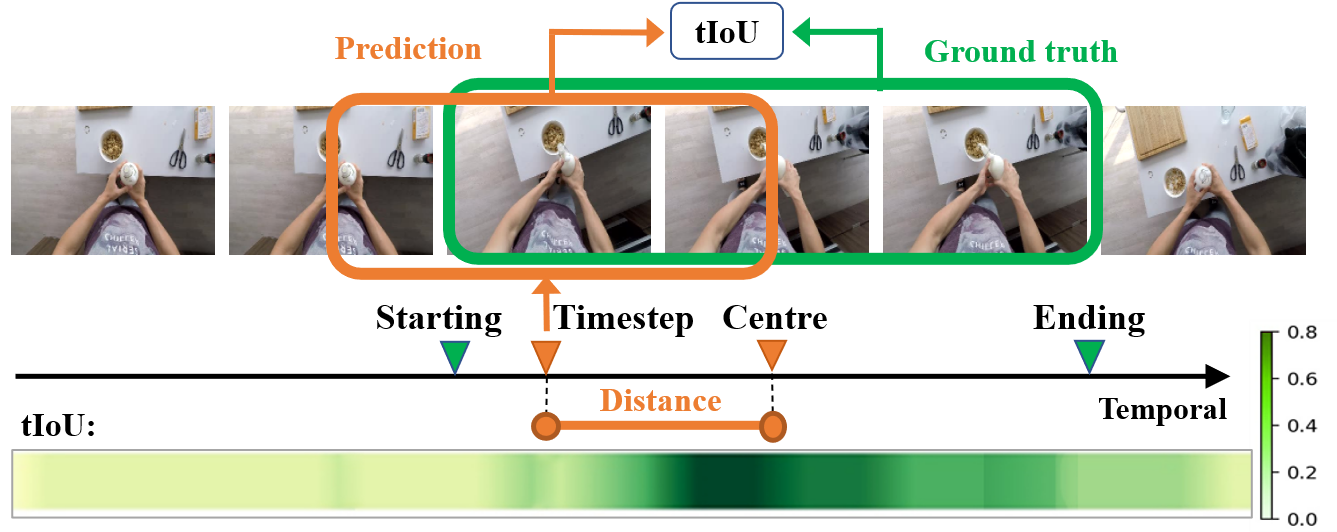}}\\
(a)&(b)
\end{tabular}
\vspace*{-10pt}
\caption{Our motivation -- (a) Sounds help detect actions both in refining the action boundaries (regression) and in identifying the action within these boundaries (classification), (b) Timesteps closer to the action center generate better proposals with high tIoU.}
\label{fig:motivation}
\end{figure}

Most recently, a few have approached egocentric action detection by modelling their long-range visual dependencies with transformers \cite{zhang2022actionformer, Tridet, wang2022refining, wang2023ego, AGT, OWL}. However, only using visual information, means a missed opportunity  to exploit  potentially meaningful aural action cues. As shown in Figure \ref{fig:motivation}(a), sound exhibits discriminating characteristics around the starting point of actions, such as `open drawer', `take spoon' and `scoop yoghurt', which can be useful for boundary regression. Also for action classification, the sound of flowing water can boost confidence in identifying an action as `turn-on tap' rather than ‘turn-off tap’, even though their visual content is similar. 
Unlike methods \cite{OWL, AVE, kazakos2021MTCN} that directly fuse audio and visual modalities at the same scale through concatenation, addition or gating modules, in this paper we learn these modalities with separate encoders and fuse their representations using a cross-modal attention mechanism at different temporal scales. This allows us to exploit sufficient audio-visual information to detect actions of various duration.

Recent one-stage anchor-free methods \cite{wang2022refining, zhang2022actionformer, Tridet} operate on egocentric videos by simultaneously predicting boundaries and action categories for each timestep. In contrast to anchor-based methods, anchor-free methods do not require pre-defined anchors to locate actions but directly generate one proposal for each timestep.
We have observed that timesteps near the centre of actions tend to produce proposals with more precise boundaries. These proposals have higher temporal Intersection-over-Union (tIoU) values with corresponding ground-truth segments. As shown in Figure \ref{fig:motivation}(b), the closer the current timestep is to the action centre, the greater the tIoU. Inspired by this observation, we introduce a  centricity head that predicts a score so as to indicate how close the current timestep is to the action centre. This score is then integral to calculating the confidence scores for ranking candidate proposals, where those with more precise boundaries will be ranked higher. 
Our approach can be incorporated into most one-stage anchor-free action detectors and achieve significant improvement.

In summary, our key contributions are as follows: (i) we introduce a framework to effectively fuse audio and visual modalities using a cross-modal attention mechanism at various temporal scales, (ii) we propose a novel centricity head to predict the degree of closeness of each frame's temporal distance to the action centre -- this boosts a proposal's confidence score and allows for the preferential selection of proposals with more precise boundaries, and (iii) we achieve state-of-the-art results on the EPIC-Kitchens-100 action detection benchmark, demonstrating the effectiveness of audio modality and the benefits of centricity in improving detection performance.

\section{Related Work}
\label{sec: related work}
\noindent\textbf{Temporal action detection --}
Current temporal action detection methods can be divided into:  
(i) two-stage methods that first generate proposals and then classify them, and (ii) one stage methods that predict boundaries and corresponding classes simultaneously. Some two-stage works generate proposals by estimating boundary probabilities \cite{BSN, BSN++, BMN, DBG} and action-ness scores \cite{SSN}.
{Many one-stage methods \cite{Turntap, SSAD, GTAN, RC3D, PBRNet} rely on pre-defined anchors to model temporal relations, which often leads to inflexibility and poor boundaries when detecting actions with various lengths. To address this, recent anchor-free methods \cite{A2Net, AFSD, zhang2022actionformer} predict the action category and offsets to the boundaries simultaneously for each timestep using parallel classification and regression heads. Then, candidate proposals constructed by these predictions are filtered to obtain the final results. Our work follows such an anchor-free pipeline.} 

{Inspired by the DETR framework \cite{DETR}, some works input relational queries \cite{shi2022react}, learned actions \cite{TadTR} or graph queries \cite{AGT} to a transformer decoder to detect actions. However, with a limited number of queries, these methods struggle to cover a large number of actions in long videos.
Alternatively, other works \cite{zhang2022actionformer, Tridet, chang2022augmented, cheng2022tallformer} use multi-scale transformer encoders \cite{zhang2022actionformer, Tridet} to model temporal dependencies for stronger video representations. For example, ActionFormer \cite{zhang2022actionformer} applies local self-attention to extract a discriminative feature pyramid, which is then used for classification and regression.  Our work falls under this workflow. } 

\noindent\textbf{Audio-visual learning --}
Sight and hearing are both vital sensory modes that assist humans in perceiving the world. This can transfer to computational approaches too to learn models from. Numerous works \cite{afouras2020self, arandjelovic2017look, aytar2016soundnet, hu2019deep, korbar2018cooperative, owens2018audio} have focused on jointly learning audio and visual representations for tasks such as action recognition \cite{kazakos2019epic, kazakos2021MTCN, Nagrani20c}, video parsing \cite{wu2021exploring, vid_parsing_2022} and event localization \cite{AVE, rao2022dual, xia2022cross}. 
Audio-visual event localization aims to classify each timestep into a limited number of categories \cite{AVE}, relying on clear audio-visual signals and without the need to predict temporal boundaries. In contrast, our action detection task aims to leverage the audio-visual representation to detect temporal boundaries for dense actions with various lengths and unclear audio cues, and then classify them into a wide range of categories.
OWL \cite{OWL} attempts different strategies for fusing audio and visual modalities, but it fuses at a single temporal scale only and classifies pre-generated proposals from \cite{BMN}, rather than detect boundaries. In \cite{lee2021cross}, the authors address this task by extracting intra-modal features, but their proposed framework is designed for simple, weakly-labelled data with sparse actions per video. Our work focuses on large-scale egocentric data comprising dense complex actions of various durations, and we propose a framework to incorporate audio-visual learning and centricity into one-stage anchor-free methods.

\section{Method}
\label{sec: method}

\begin{figure}[t!]
	\centering
	\includegraphics[width=0.95\linewidth]{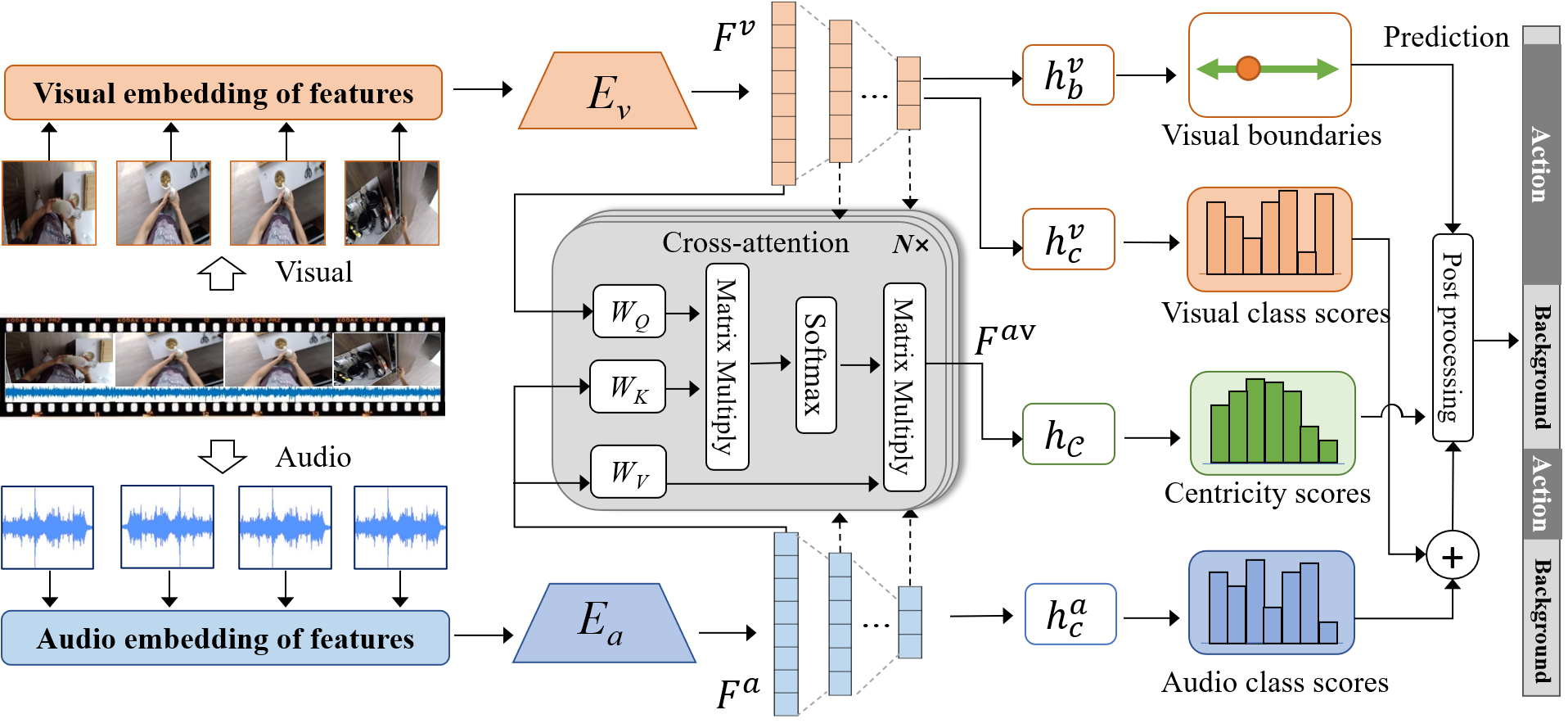}
	\vspace*{-5pt}
    \caption{An overview of our architecture -- Given an untrimmed video, audio and visual features are extracted from video clips, then fed into encoders $E_{a}$ and $E_{v}$ for generating audio $F^{a}$ and visual $F^{v}$ feature pyramids. These are fused using cross-attention across {$N$} temporal scales to build the audio-visual representation $F^{av}$. This representation is passed to the centricity head $h_{{\mathcal{C}}}$ for centricity scores, and to prediction heads $h^{v}_{b}$, $h^{v}_{c}$, and $h^{a}_{c}$ for boundary, classification, and audio scores respectively. Finally, predicted scores and boundaries are used to construct candidate proposals, which are then filtered to obtain the final predictions.}
\label{fig:arch}
\end{figure}

We propose a novel framework for temporal action detection, rooted in audio-visual data, which can be incorporated into one-stage anchor-free pipelines \cite{zhang2022actionformer, AFSD, wang2022refining, Tridet} (see Figure \ref{fig:arch}). 
\noindent Similar to such temporal action detection works, 
we define the problem as follows. Given an untrimmed video, we extract features for the video and audio modalities and then process them using transformer encoders to obtain the visual and audio representation sequences $F^{v} =  \left\{f^{v}_{t}\right\}^{T}_{t=1}$ and $F^{a} =  \left\{f^{a}_{t}\right\}^{T}_{t=1}$, respectively. Based on these, our approach is to learn to predict a set of possible action instances $\Phi=\left\{(s,e,{\alpha})_m \right\}^M_{m=1}$, where $s$ and $e$ represent the starting and ending boundaries of an action, and $\alpha$ represents the predicted action class.

\begin{figure} 
\begin{tabular}{ccc}
\bmvaHangBox{\includegraphics[width=0.31\linewidth]{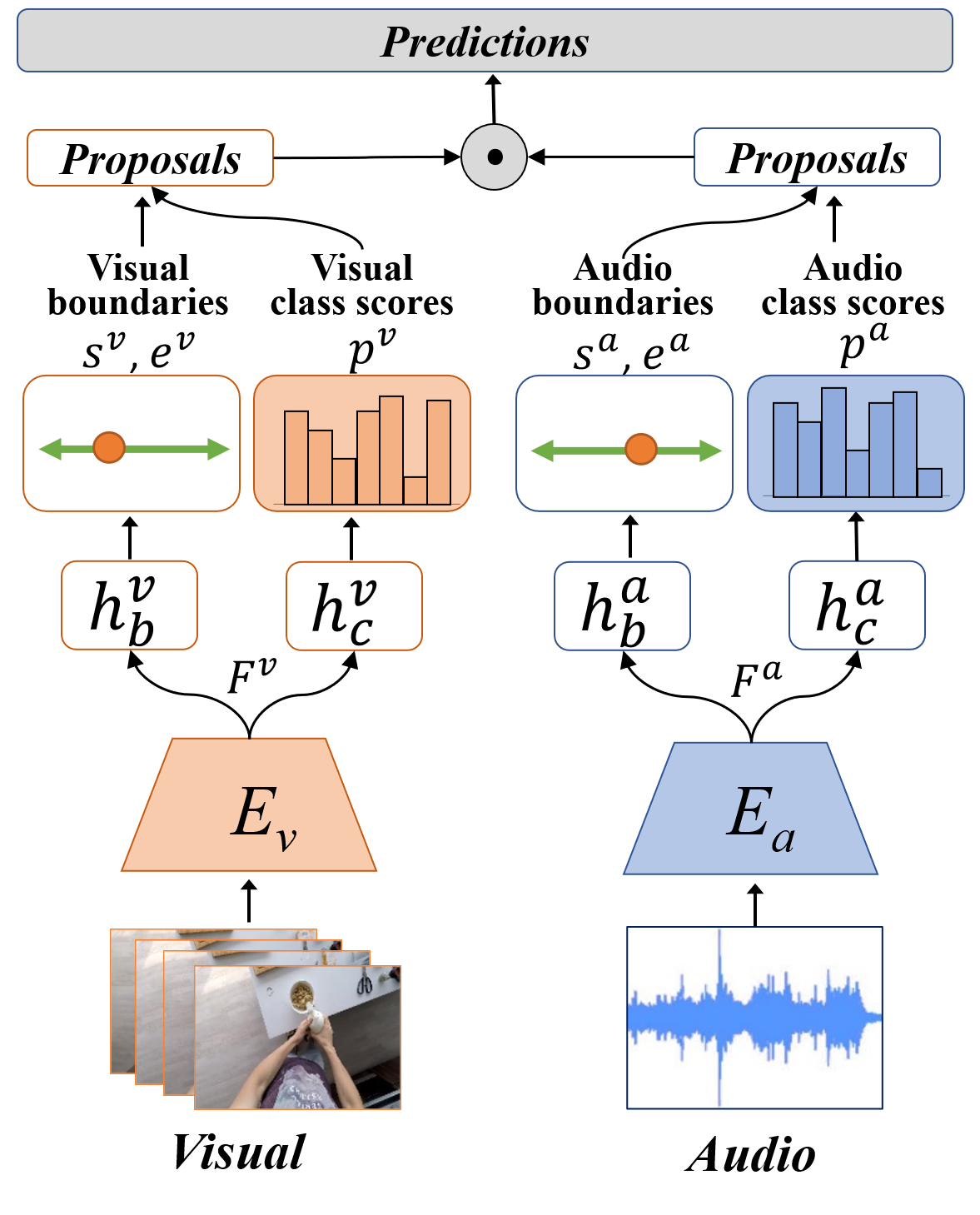}}&
\bmvaHangBox{\includegraphics[width=0.30\linewidth]{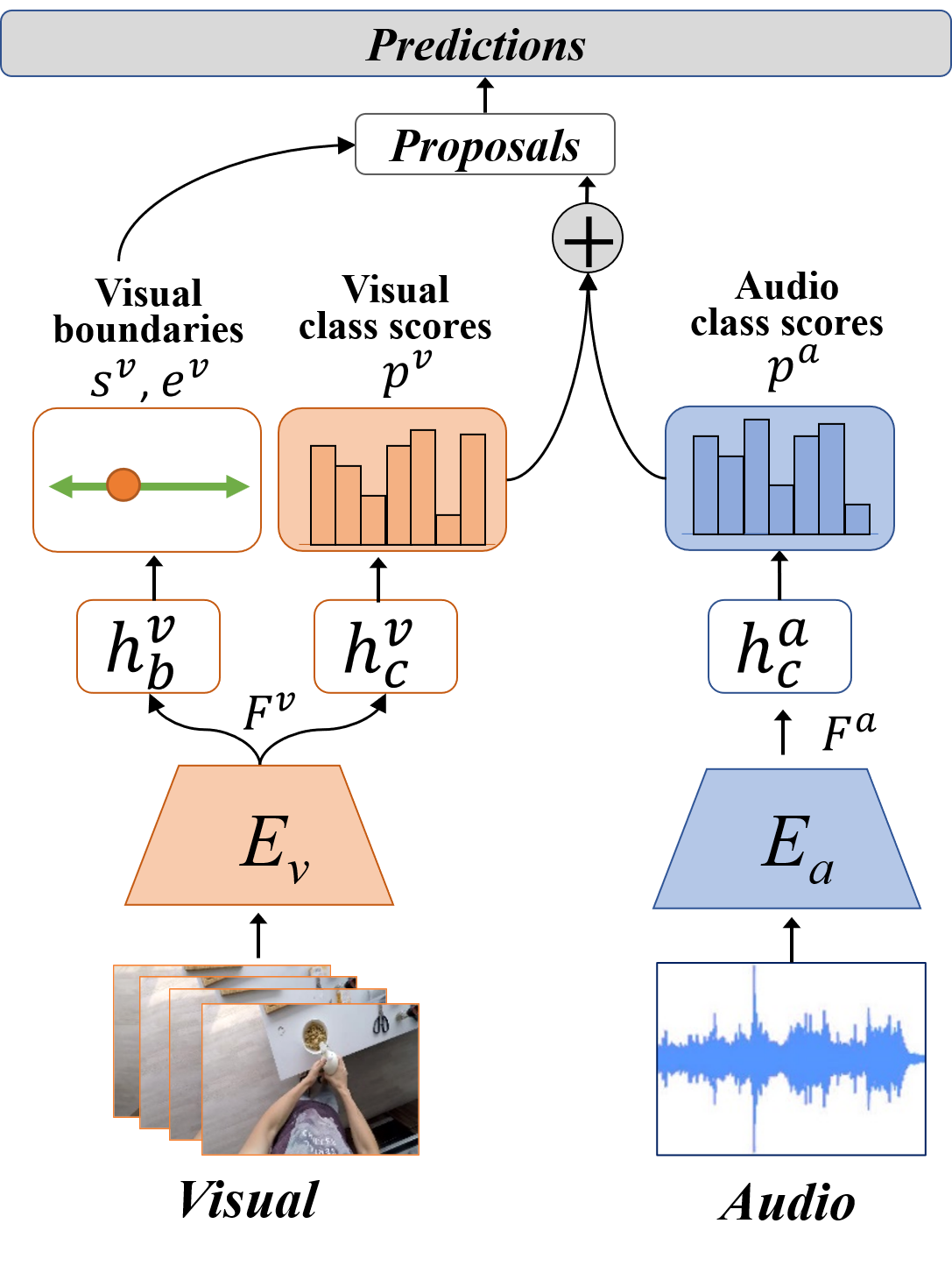}}&
\bmvaHangBox{\includegraphics[width=0.3025\linewidth]{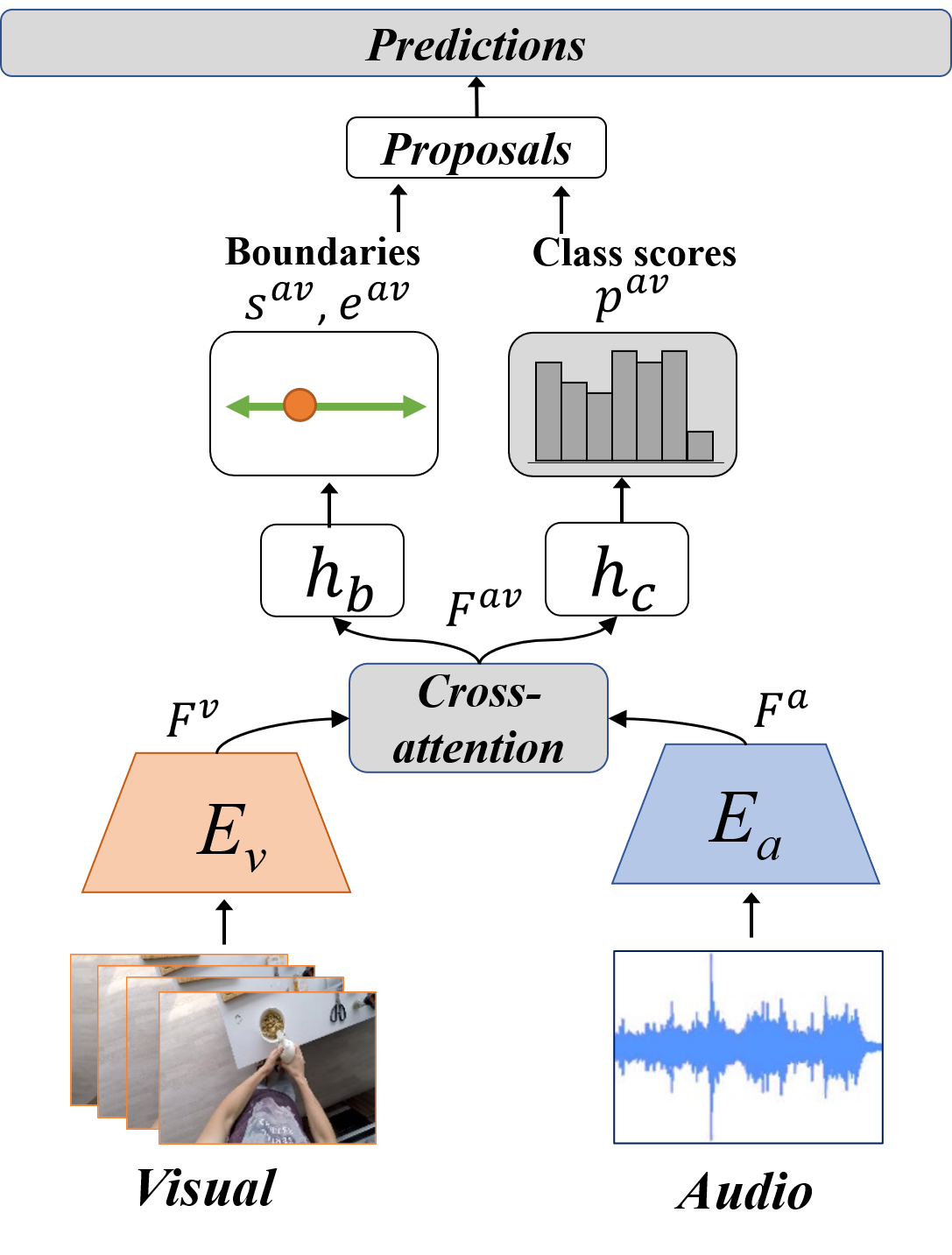}}\\
(a)&(b)&(c)
\end{tabular}
\caption{Three strategies for fusing modalities -- (a) Proposal fusion: visual and audio modalities are separately fed into their respective streams and generate corresponding sets of proposals, which are directly {concatenated} to obtain the final set of proposals, (b) Classification scores fusion: classification scores from both  modalities are combined alongside boundaries from the video  to obtain the final set of proposals, (c) Feature pyramid fusion: feature pyramids from the two modalities are fused through cross-attention and fed into parallel heads to predict boundaries and class scores, which then construct the set of proposals.}
\label{fig:diffusion}
\end{figure}

\subsection{Audio-visual Fusion}
\label{subsec:fusion}
In this section, we explore three different strategies to effectively utilise the audio modality and combine it with visual information to improve action detection performance. 

\noindent\textbf{Proposal fusion --}
In this strategy (see Figure \ref{fig:diffusion} (a)), at first the visual and audio representations $(F^{v},F^{a})$ are produced by encoders $E_{v}$ and $E_{a}$ respectively.  Classification heads $h^{v}_{c}$ and $h^{a}_{c}$ and regression heads $h^{v}_{b}$ and $h^{a}_{b}$ are then used to predict the classification scores $(p_{t}^{v}$,$p_{t}^{a})$ and boundaries $(s^{v}_{t}$,$e^{v}_{t})$ and $(s^{a}_{t}$,$e^{a}_{t})$ 
for the visual and audio modalities, respectively. 
Thus, we can obtain a set of candidate proposals for the visual modality $\Phi_{v}=\left\{(s^{v}_{t},\ e^{v}_{t},\ p^{v}_{t}) \right\}^T_{t=1}$ and similarly, a set of candidate proposals for the audio modality $\Phi_{a}=\left\{(s^{a}_{t},\ e^{a}_{t},\ p^{a}_{t}) \right\}^T_{t=1}$. Then, we {concatenate} these two sets as 
$\Phi_{{o}} = \left\{(s^{v}_{t},\ e^{v}_{t},\ p^{v}_{t}), (s^{a}_{t},\ e^{a}_{t},\ p^{a}_{t}) \right\}^T_{t=1}$.

\noindent\textbf{Classification scores fusion --}
{Although sounds can be associated with actions for classification purposes, the duration of an action does not necessarily correspond to its audio start and end as recently shown in~\cite{EPICSOUNDS2023}. Thus, we discard the audio boundaries, integrate the classification scores from both visual and audio modalities, and then use them  along with the visual boundaries to generate proposals.}

We use an approach similar to \cite{OWL} to fuse visual and audio classification scores. Specifically, as shown in Figure \ref{fig:diffusion}(b), based on $F^{v}$ and $F^{a}$,  the visual classification head $h^{v}_{c}$, audio classification head $h^{a}_{c}$ and visual boundary head $h^{v}_{b}$ predict  scores $p_{t}^{v}$ and $p_{t}^{a}$, and frame boundaries $s^{v}_{t}$ and $e^{v}_{t}$, respectively.
We fuse the classification scores $p_{t}^{v}$ and $p_{t}^{a}$ by simple addition and combine them with the visual boundaries $s^{v}_{t}$ and $e^{v}_{t}$ to construct the set of fused candidate proposals $\Phi_{c}$, such that
$\Phi_{c} = \left\{(s^{v}_{t},\ e^{v}_{t},\ p^{v}_{t}+p^{a}_{t})) \right\}^T_{t=1}$.

\noindent\textbf{Feature pyramid fusion --}
{In this strategy (see Figure \ref{fig:diffusion}(c)), $F^{v}$ and $F^{a}$ are put through a cross-attention mechanism \cite{lee2021cross, OWL} (see also grey box of Figure \ref{fig:arch}) to model their inter-modal {dependencies} which then results in a single representation vector $F^{av}$. }

Firstly, $F^{v}$ and $F^{a}$ are projected into query  $Q = W_{Q}F^{v}$, key $K = W_{K}F^{a}$, and value $V = W_{V}F^{a}$,
where $F^{v}$ serves as a query input and $F^{a}$ serves as key and value inputs. $W_{Q}$, $W_{K}$ and $W_{V} \in \mathbb{R}^{d \times d}$ are learnable weight matrices, where $d$ is the embedding dimension. Next, we calculate the audio-visual representation vector 
\begin{equation}
\label{attn feat}
F^{av} = softmax(\frac{QK^{T}}{\sqrt{d}})V~,
\end{equation}
which is then  fed into a classification head {$h_{c}$} and a regression head {$h_{b}$} to obtain the classification scores $p^{av}_{t}$ and the boundaries $s^{av}_{t}$, $e^{av}_{t}$ for each timestep. Therefore, the set of candidate proposals is
$\Phi_{f} = \left\{(s^{av}_{t},\ e^{av}_{t},\ p^{av}_{t})) \right\}^T_{t=1}$.

We ablate these three strategies in Sec. \ref{abaltion}. Based on the ablations, we chose feature pyramid fusion to generate audio-visual representations {across $N$ temporal pyramid scales} for assessment by the centricity head to predict corresponding scores (see Sec. \ref{centerness} next). We also selected the classification scores fusion approach to generate stronger audio-visual classification scores to predict action categories and calculate confidence scores.

\subsection{Audio-visual Centricity Head}
\label{centerness}
\begin{figure}[h!]
\begin{tabular}{cc}
\bmvaHangBox{\includegraphics[width=0.34\linewidth]{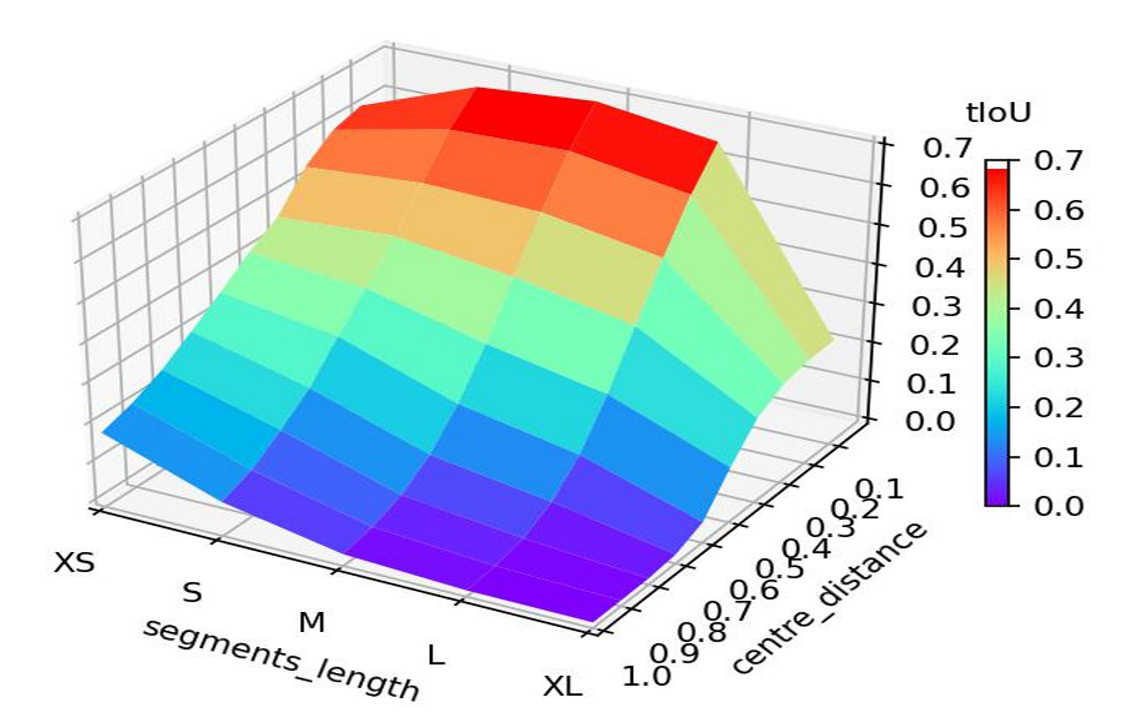}}&\hspace{-4mm}
\bmvaHangBox{\includegraphics[width=0.59\linewidth]{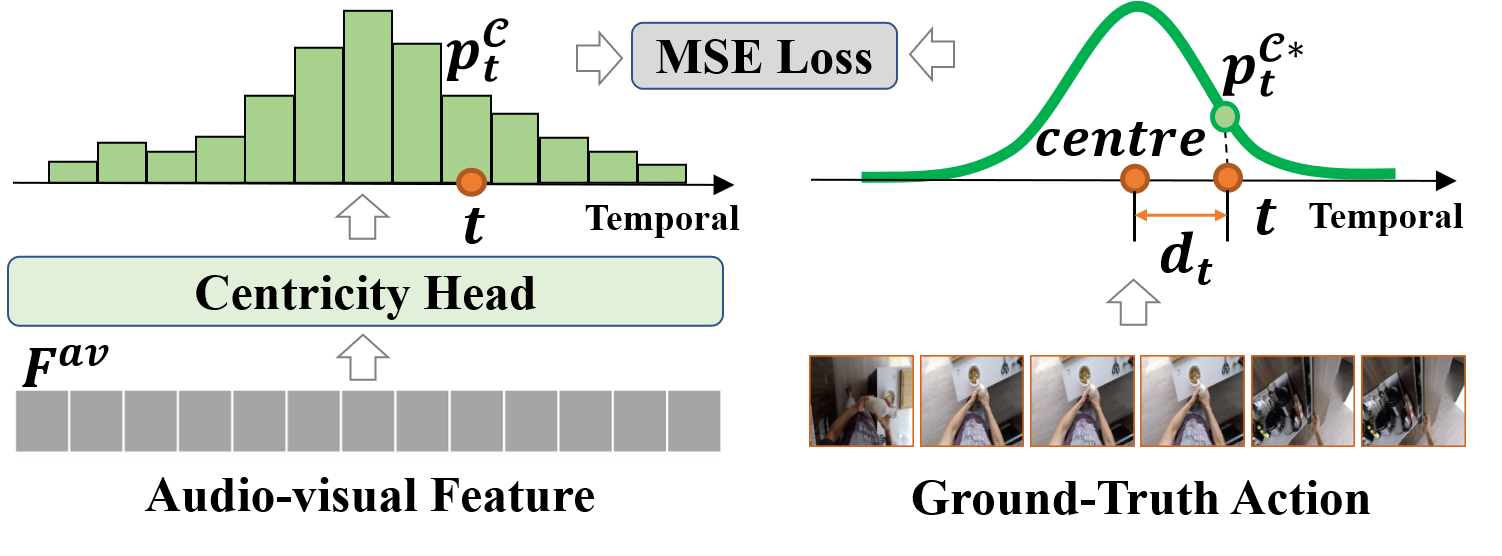}}\\
(a)&(b)
\end{tabular}
\caption{The illustration of centricity -- {(a) The tIoUs between ground-truth and predictions (from RAB \cite{wang2022refining}) is plotted across various centre distances. Actions are divided into five groups based on segment lengths (in seconds): XS (0, 2], S (2, 4], M (4, 6], L (6, 8], and XL (8, inf).  (b) The centricity head takes in the audio-visual feature $F^{av}$ and produces the centricity scores $p_{t}^{{\mathcal{C}}}$. The ground-truth centricity score $p_{t}^{{\mathcal{C}}*}$ is calculated based on the relative distance $d_{t}$ between the time step $t$ and the action center.}}

\label{fig:centricity}
\end{figure}

We investigated the relationship between the distance of a timestep from the action centre and {the tIoU value between its generated proposal and the ground truth}. As shown in Figure \ref{fig:centricity}(a), as a timestep gets closer to the action center, its generated proposal has a higher tIoU value.
This indicates that timesteps around the centre of an action can generate proposals with more reliable action boundaries. 
Thus, we propose a simple, yet effective, centricity head based on the audio-visual representation {$F^{av}$} to estimate how close the timestep $t$ is to the centre of the action (as shown in Figure \ref{fig:centricity} (b)).
The centricity head consists of three 1D conv layers with layer normalization and a ReLU activation function.

\noindent\textbf{Label assignment --} We require the centricity scores $p_{t}^{\mathcal{C}*}$ calculated from ground-truth data as supervision signals for training. 
For each timestep $t$, we consider the relative distance $d_{t}$ between the current timestep $t$ and the centre of the corresponding ground-truth action to map the training labels of centricity scores
\begin{equation}
  p_{t}^{{\mathcal{C}}*}= \text{exp} \left({-(d_{t})^{2}/2\sigma^{2}} \right) ~,
  \label{eq:gau_label}
\end{equation}
where $\sigma$ is a scaling hyperparameter which defines that the closer a timestep is to the action centre, the higher the centricity score. This has previously been explored to predict  boundary confidences~\cite{Gaussian2016, wang2022refining}, but not specifically for centricity use cases. The centricity scores are normalized to a range of 0 to 1.

\noindent\textbf{Training --} We optimize the loss between the ground-truth $p_{t}^{\mathcal{C}*}$ and the predicted centricity scores $p_{t}^{\mathcal{C}}$ using Mean Square Error (MSE) loss as 
\begin{equation}
\begin{aligned}
{L}_{{\mathcal{C}}}=\frac{1}{T^{'}}\sum_{t}^{T^{'}} (p_{t}^{{\mathcal{C}}*}-p_{t}^{{\mathcal{C}}})^{2}
\end{aligned} ~,
\end{equation}
where $T^{'}$ is the total number of timesteps used for training from all scales of the audio-visual feature pyramid.
Our method can be integrated into any one-stage anchor-free frameworks and trained in an end-to-end manner. 
The total loss is 
$L_{total} = L_{g} + \lambda_{1} L_{c} + \lambda_{2} L_{b} + \lambda_{3} L_{\mathcal{C}}~$,
where $L_{g}$ and $L_{c}$ are losses for regression \cite{iouloss} and classification \cite{focal} and are the same as in \cite{zhang2022actionformer, Tridet, wang2022refining}. $L_{b}$ is the boundary confidence loss from \cite{wang2022refining}. $\lambda_{1}$, $\lambda_{2}$ and $\lambda_{3}$ denote the loss balancing weights, and $\lambda_{3}$ is set to 0 when the baseline is ActionFormer \cite{zhang2022actionformer} or TriDet \cite{Tridet}.

\subsection{Post-processing}
\label{post-processing}
{For each timestep $t$, the network produces the visual and audio classification scores ${p}_{t}^{v}$, ${p}_{t}^{a}$, the corresponding action class label $\alpha$, a centricity score ${p}_{t}^{{\mathcal{C}}}$, and a pair of starting and ending boundaries ${s_t}$, ${e_t}$ with their corresponding boundary confidences ${p}_{{s_t}}^{s}$ and ${p}_{{e_t}}^{e}$ (with these confidences computed as in \cite{wang2022refining}). 
The final confidence score for timestep $t$ is then a weighted combination of the learnt knowledge, i.e. 
\begin{equation}
\label{conf score equation}
{{\mathcal{S}}} = {p}_{t}^{v} + {\tau} {p}_{t}^{a} + \beta {p}_{t}^{{\mathcal{C}}}  + \gamma (p^{s}_{s_{t}} + p^{e}_{e_{t}}) ,
\end{equation}
where {$\tau$}, $\beta$ and $\gamma$ are fusion weights, and $\gamma$ is set to 0 when the baseline is ActionFormer\cite{zhang2022actionformer} or TriDet \cite{Tridet}. Finally, we follow  standard practice \cite{BSN, BMN, BSN++, zhang2022actionformer} to rank these candidate actions based on the final confidence score $\mathcal{S}$ and filter them using Soft-NMS \cite{softnms} to obtain a final set of $M$ predictions $\Phi=\left\{(s,e,\alpha)_m \right\}^M_{m=1}$.
}

\section{Experiments}
\noindent {\bf Dataset --}
We conduct experiments on EPIC-Kitchens-100 \cite{EPIC100},  a large-scale audio-visual dataset that contains 700  unscripted videos with 97 verb and 300 noun classes. On average, there are 128 action instances per video, with significant overlap.

\noindent {\bf Evaluation metric --}
We use mean Average Precision (mAP) for verb, noun and action tasks at various IoU thresholds \{0.1, 0.2, 0.3, 0.4, 0.5\} to evaluate our method against others.

\noindent {\bf Baselines --}
Our approach is integrated with three one-stage anchor-free approaches \cite{zhang2022actionformer, Tridet, wang2022refining}. ActionFormer \cite{zhang2022actionformer} is chosen as it is a pioneering anchor-free work that models long-range temporal dependencies using the Transformer for action detection. TriDet \cite{Tridet} extends this by incorporating scalable-granularity perception layers and a Trident head to regress boundaries. Finally, Wang et al. \cite{wang2022refining} (hereafter RAB) introduces a method to estimate boundary confidences through  Gaussian scaling.

{\noindent {\bf Implementation details --}
We compare our approach against state-of-the-art (SOTA) methods \cite{EPIC100, BMN, OWL, TADA, zhang2022actionformer, wang2022refining, Tridet} for temporal action detection. For a fair comparison, we employ the same visual features as \cite{zhang2022actionformer, Tridet, wang2022refining}, extracted from an action recognition model \cite{EPIC100} that is pre-trained with the SlowFast \cite{slowfast} network on EPIC-Kitchens-100. To obtain features with a dimension of 1x2304, we have a window size of 32 and a stride of 16 frames.
For audio features, we generate 512×128 spectrograms using a window size of 2.6ms and a stride of 1.3ms. These spectrograms are then fed into a SlowFast audio recognition model \cite{kazakos2021slow}, with features extracted after the average pooling layer with a dimension 1x2304.}

Again, following \cite{zhang2022actionformer, Tridet, wang2022refining}, the feature pyramid generated by the transformer encoder has {$N$} = 6 levels, with a level scaling factor of 2.
For training, we use one Nvidia P100 GPU. We crop the video features with various lengths to 2304. The loss balancing weights in Sec. \ref{centerness} are  $\lambda_{1} = 1$, $\lambda_{2} = 0.5$, and $\lambda_{3} = 1.7$. The weight ratio of the classification loss between verb and noun is set to 2:3.
The scaling hyperparameter in Eq. (\ref{eq:gau_label}) is set to $\sigma = 1.7$. During the inference stage, the confidence score weights in Sect. \ref{post-processing} are assigned as $\tau = 0.2$, $\beta = 1$, and $\gamma = 0.7$. For the multi-task classification, we select the top 11 verb and the top 33 noun predictions to combine the candidate actions.

\noindent\textbf{Main results --}
Table \ref{tab:results_epic} shows that the proposed method outperforms recent SOTA approaches \cite{EPIC100, BMN, OWL, TADA, zhang2022actionformer, wang2022refining, Tridet} on the EPIC-Kitchens-100 action detection benchmark, and achieves significant improvements when added to existing SOTA one-stage multi-scale methods \cite{zhang2022actionformer, wang2022refining, Tridet}. 
{ActionFormer \cite{zhang2022actionformer} and TriDet \cite{Tridet} train different models for verb and noun detection and do not detect actions. Instead, we train one model and add a multi-task classification head to predict their results for the action task. It can be seen that our enhanced proposed method with audio fusion and centricity improves performance in every metric. In  action detection, mAP improves by $1.35\%$ and $0.97\%$, respectively.  RAB~\cite{wang2022refining} also performs well on egocentric data and follows the same anchor-free pipeline as ours. The improvement achieved on RAB\cite{wang2022refining} was $1.32\%$ and is also the best  result  amongst the baselines.}
\begin{table*}[t!]
\scriptsize
\begin{center}
\begin{tabular}{|l|c|c|c|ccc|}
\hline
\multicolumn{1}{|c}{\multirow {2}{*}{\bf Method}} &\multicolumn{1}{|c}{\multirow {2}{*}{\bf Venue}} 
&\multicolumn{1}{|c}{\multirow {2}{*}{\bf Feature}} 
&\multicolumn{1}{|c}{\multirow {2}{*}{\bf Audio}} &\multicolumn{3}{|c|}{\bf Avg. mAP@task}\\
\cline{5-7}
\multicolumn{1}{|c}{}&\multicolumn{1}{|c}{}&\multicolumn{1}{|c}{}&\multicolumn{1}{|c}{}&\multicolumn{1}{|c}{Verb} & Noun & Action \\ 
\hline\hline
\multirow{1}{*}{BMN \cite{BMN,EPIC100}} &IJCV 2022 &SF \cite{slowfast} &$\times$ &8.36 &6.53 &5.21 \\ 
\multirow{1}{*}{OWL \cite{OWL}} &ArXiv 2022 &SF \cite{slowfast} &\checkmark &11.47 &12.63 &8.35 \\
\multirow{1}{*}{BMN+TSN \cite{BMN, TADA}} &ICLR 2022 &TSN \cite{TSN} &$\times$ &13.47 &12.37 &9.71 \\
\multirow{1}{*}{BMN+TAda2D \cite{BMN, TADA}} &ICLR 2022 &TAda2D \cite{TADA} &$\times$ &16.78 &17.39 &13.18\\
\hline
\multirow{1}{*}{ActionFormer~\cite{zhang2022actionformer}$^*$} &ECCV 2022 &SF \cite{slowfast} &$\times$ &20.45  &20.90  &16.63\\
\multirow{1}{*}{ActionFormer~\cite{zhang2022actionformer}$^*$ + Ours} &- &SF \cite{slowfast} &\checkmark &20.48 &22.41 &17.98  \\ 
\hline
\multirow{1}{*}{TriDet~\cite{Tridet}$^*$} &CVPR 2023 &SF \cite{slowfast} &$\times$ &20.87  &21.04  &17.21 \\
\multirow{1}{*}{TriDet~\cite{Tridet}$^*$ + Ours} &- &SF \cite{slowfast} &\checkmark &\textbf{21.94}  &22.86  &18.18 \\ 
\hline
\multirow{1}{*}{RAB \cite{wang2022refining}} &AVSS 2022 &SF \cite{slowfast} &$\times$ &20.71 &20.53 &17.18\\
\multirow{1}{*}{RAB \cite{wang2022refining} + Ours} &- &SF \cite{slowfast} &\checkmark &21.10 &\textbf{23.08} &\textbf{18.50}  \\
\hline
\end{tabular}
\end{center}
\caption{Comparative results on the EPIC-Kitchens-100 action detection validation set -- $^*$Actionformer and TriDet only provide results on verb and noun detection, hence we produce action results by modifying them with a multi-task action classification head \cite{wang2022refining}.
}
\label{tab:results_epic}
\end{table*}

\begin{figure}[t!]
	\centering
	\includegraphics[width=0.90\linewidth]{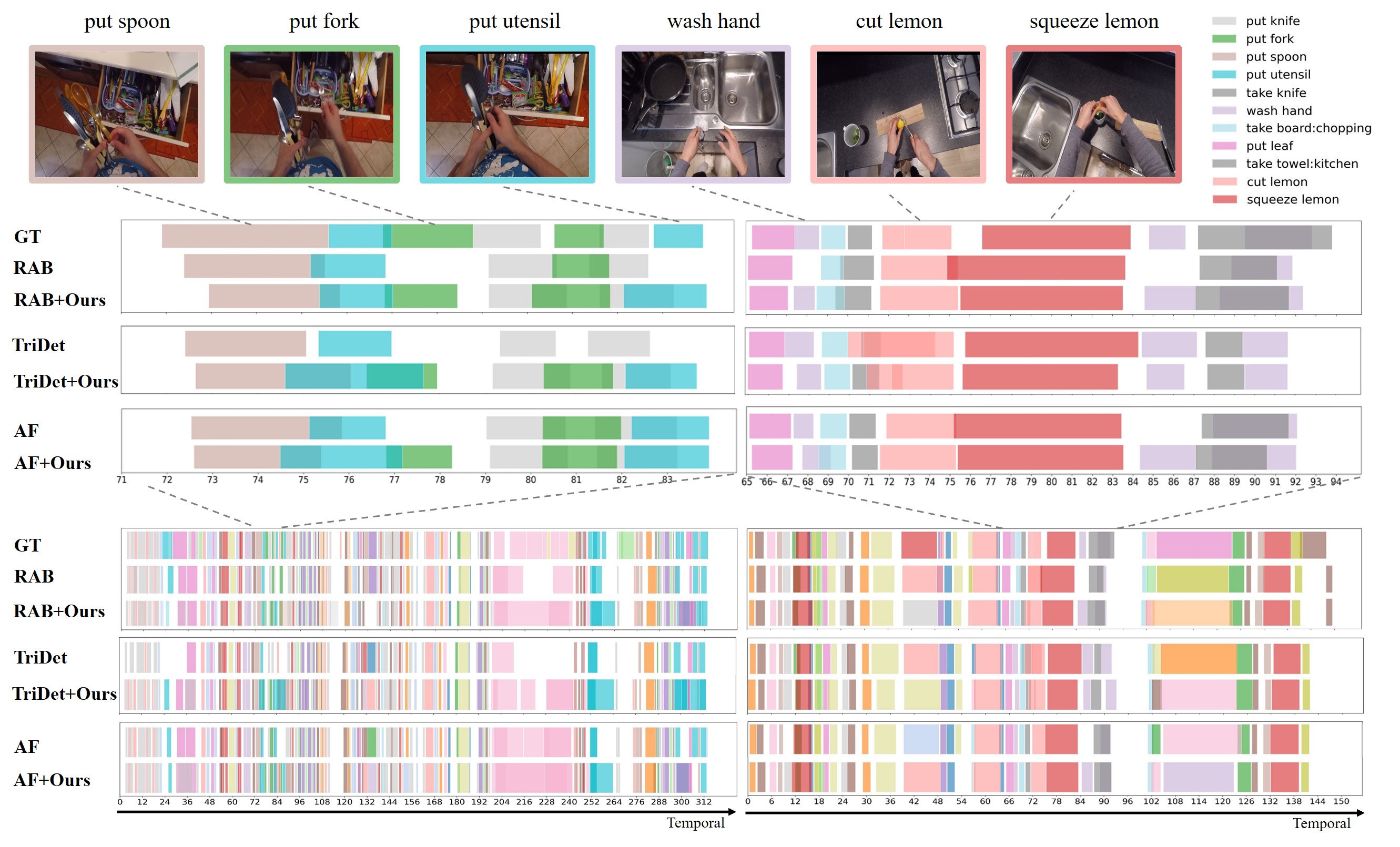}
	\vspace*{-10pt}
    \caption{Qualitative results on the EPIC-Kitchens-100 action detection dataset -- The top row shows the visual content of selected frames. The middle seven lines display the ground-truth (GT), the predictions of RAB \cite{wang2022refining}, RAB \cite{wang2022refining}+Ours, TriDet \cite{Tridet}, TriDet \cite{Tridet}+Ours, AF \cite{zhang2022actionformer}, AF \cite{zhang2022actionformer}+ours for a zoomed-in region.  The bottom seven lines represent the whole video sequence. These results demonstrate the effectiveness of our method in accurately detecting dense actions.}
	\label{fig: qualitative}
\end{figure}
\noindent\textbf{Qualitative results --}
Qualitative plots of `RAB \cite{wang2022refining}+Ours' on the EPIC-Kitchens-100 action detection validation dataset are shown in Figure \ref{fig: qualitative}. The bottom two lines showcase the model's ability to detect dense actions with different classes and durations in a video, demonstrating that our approach can effectively utilise the audio modality to learn discriminative representations. 
The middle three lines show a zoomed-in, detailed look where it is easier to see that our method better deals with challenging actions, e.g. see action `put fork' in the $77^{th}$ second of the first video and the action `wash hand' in the $68^{th}$ second of the second video that were missed by the baseline model. This indicates that our centricity head enhances the confidence scores for actions with more precise boundaries, resulting in their preferential ranking and selection during the Soft-NMS processing.

\subsection{Ablations}  \label{abaltion}
{All our ablations use RAB \cite{wang2022refining} as baseline and are performed on EPIC-Kitchens-100 {validation dataset}~\cite{EPIC100}.}

\noindent {\bf Components ablation --} We ablate the contributions of our two main components, audio-visual fusion and centricity head, as seen in Table \ref{tab:Component}. Both components demonstrate notable improvements in the performance of the baseline methods  \cite{zhang2022actionformer, Tridet, wang2022refining}, particularly when they are engaged, singularly or in combination, for action detection.

\begin{table}[t]
\scriptsize
\begin{center}
\begin{tabular}{|l|c|c|ccc|}
\hline
\multicolumn{1}{|c}{\multirow {2}{*}{\bf Baseline
}} &\multicolumn{1}{|c}{\multirow {2}{*}{\bf Audio}} &\multicolumn{1}{|c}{\multirow {2}{*}{\bf \makecell{Centri-\\city}}} &\multicolumn{3}{|c|}{\bf Avg. mAP@task}\\
\cline{4-6}
\multicolumn{1}{|c}{}&\multicolumn{1}{|c}{}&\multicolumn{1}{|c}{}&\multicolumn{1}{|c}{Verb} & Noun & Action \\ 
\hline\hline
\multirow{1}{*}{ActionFormer \cite{zhang2022actionformer}} &$\times$ &$\times$  &20.45  &20.90  &16.63  \\
\multirow{1}{*}{ActionFormer \cite{zhang2022actionformer} + Ours} &\checkmark &$\times$ &19.75 &21.46 &17.22  \\
\multirow{1}{*}{ActionFormer \cite{zhang2022actionformer} + Ours} &$\times$ &\checkmark&\textbf{21.50} &22.25 &17.60  \\
\multirow{1}{*}{ActionFormer \cite{zhang2022actionformer} + Ours} &\checkmark &\checkmark &20.48 &\textbf{22.41} &\textbf{17.98}  \\
\hline
\multirow{1}{*}{TriDet~\cite{Tridet}} &$\times$ &$\times$ &20.87  &21.04  &17.21 \\
\multirow{1}{*}{TriDet~\cite{Tridet} + Ours}&\checkmark &$\times$ &21.07  &21.75  &17.61 \\
\multirow{1}{*}{TriDet~\cite{Tridet} + Ours}&$\times$ &\checkmark &21.65  &21.23  &17.42 \\
\multirow{1}{*}{TriDet~\cite{Tridet} + Ours}&\checkmark &\checkmark &\textbf{21.94}  &\textbf{22.86}  &\textbf{18.18} \\
\hline
\multirow{1}{*}{RAB \cite{wang2022refining}} &$\times$ &$\times$ &20.71 &20.53 &17.18 \\
\multirow{1}{*}{RAB \cite{wang2022refining} + Ours}&\checkmark &$\times$ &20.93 &21.84 &17.88 \\
\multirow{1}{*}{RAB \cite{wang2022refining} + Ours}&$\times$ &\checkmark &\textbf{21.47} &21.43 &17.69\\
\multirow{1}{*}{RAB \cite{wang2022refining} + Ours}&\checkmark &\checkmark &21.10 &\textbf{23.08} &\textbf{18.50}  \\
\hline
\end{tabular}
\end{center}
\vspace*{-3pt}
\caption{Components analysis on the EPIC-Kitchens-100 action detection validation set.
}
\label{tab:Component}
\end{table}

\begin{table}[t!]
\scriptsize
\begin{minipage}[h]{0.45\textwidth}
\vspace{-0.2cm}
\begin{center}
\setlength{\tabcolsep}{1.5mm}
{
\begin{tabular}{|c|ccc|}
\hline
\multicolumn{1}{|c|}{\multirow {2}{*}{\bf \makecell{Classification \\Weight $\lambda_{1}$}}} &\multicolumn{3}{c|}{\bf Avg. mAP@task}\\
\cline{2-4}
\multicolumn{1}{|c|}{}&\multicolumn{1}{c}{Verb} &Noun &Action 
\\ 
\hline\hline
0.5 &22.04 &22.70 &18.22
 \\
1 &21.10 &\textbf{23.08} &\textbf{18.50}
\\
2 &22.13 &22.52 &18.01
\\
4 &\textbf{22.36} &22.11 &18.09
 \\
6  &22.21 &22.75 &17.83
\\
8  &22.20 &22.92 &17.83
\\
\hline
\end{tabular}}
\end{center}
\caption{Ablation on varying classification weight  $\lambda_{1}$.}
\label{Ablations Tab: classification}
\end{minipage}
\hspace{0.01\textwidth}
\makeatletter\def\@captype{figure}\makeatother
\begin{minipage}[h]{0.50\textwidth}
\vspace{-0.2cm}
\centering
\includegraphics[width=0.69\linewidth]{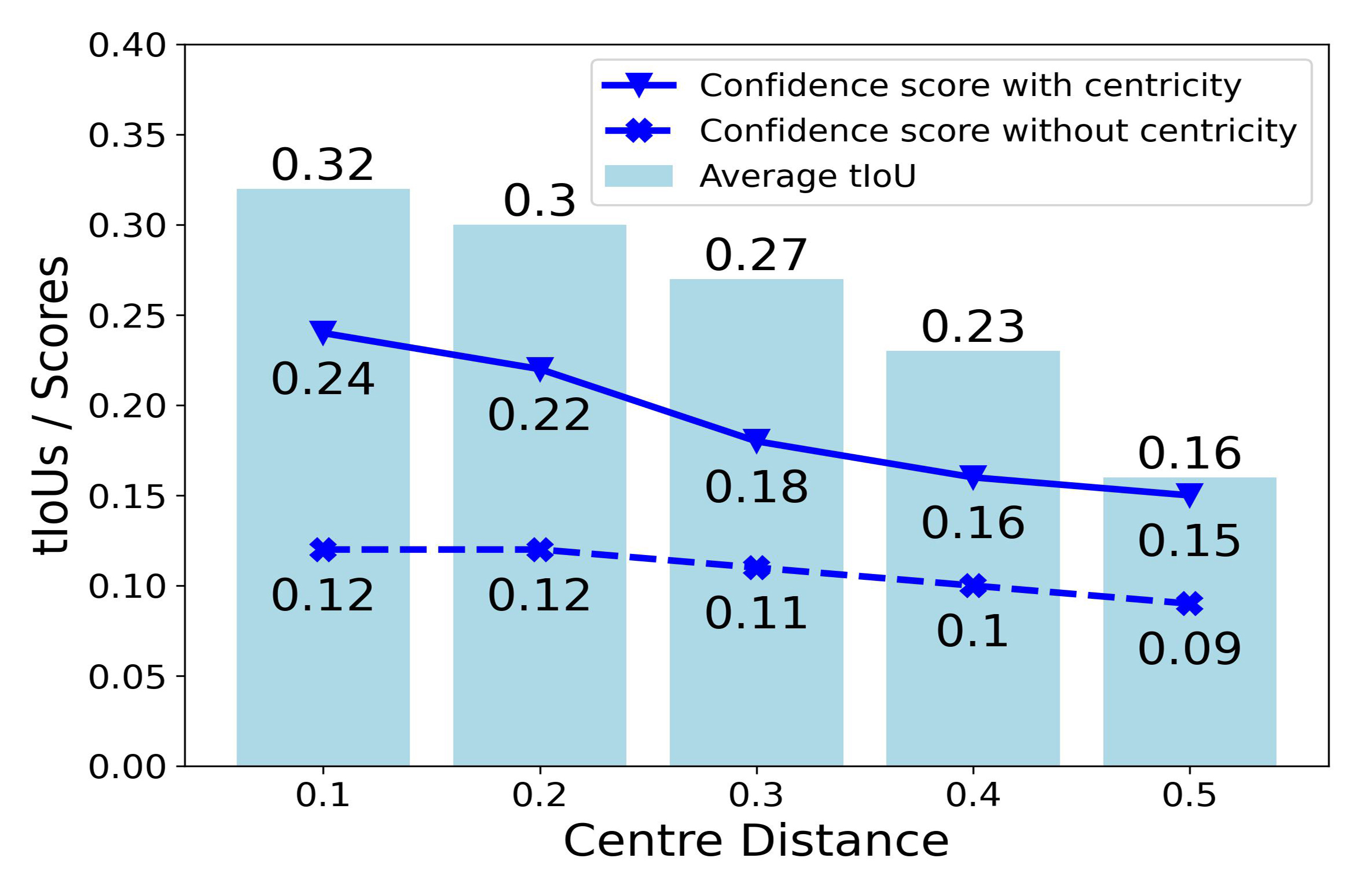}
\vspace*{-8pt}
\caption{Effect of centricity on confidence scores (see text for details).}
\label{fig:effect of centricity}
\end{minipage}
\vspace*{-2pt}
\end{table}

{\noindent {\bf Loss function weights --}
We train our network in an end-to-end manner by minimizing the total loss function in Sec. \ref{centerness}, where we assign three weights to balance various losses. Table \ref{Ablations Tab: classification} shows that $\lambda_{1} = 1$ is the best value for performing the action task for the classification loss $L_c$ weight, across the improvements made to the baseline. For the boundary confidence loss  $L_b$ weight $\lambda_{2}$, we follow the baseline RAB's recommended setting \cite{wang2022refining}. Finally, 
Table \ref{Ablations Tab: centricity weight} demonstrates that our results are relatively stable when varying centricity loss $L_\mathcal{C}$ weight $\lambda_{3}$ between $0.5-2$, with the best action {detection} result at $\lambda_{3}=1.7$.
}

{\noindent {\bf The effect of centricity on confidence score --}} {Figure \ref{fig:effect of centricity} shows that as the centre distance increases, the average tIoU values (blue bars) between the ground truth and the proposals generated by timesteps at the centre distance present a notable decreasing trend ($\downarrow$0.16). However, the original confidence scores (dashed blue line) only slightly decreases ($\downarrow$0.03). 
Adding centricity into the confidence scores (solid blue line) responds better to the expected trend ($\downarrow$0.09). Thus, proposals with more accurate boundaries (higher tIoU values) rank higher based on their confidence scores when centricity is incorporated.}

\noindent {\bf Audio-visual fusion strategies --} The three strategies for fusing audio and visual modalities (see Section \ref{subsec:fusion}) are compared in Table \ref{tab:diff fusion}. The proposal fusion strategy has the lowest overall action {detection} performance due to the audio stream's proposals having less precise boundaries. The audio-visual fusion of classification scores strategy improves on Visual-only through multiplication ($\uparrow$0.19\%) or addition ($\uparrow$0.51\%). For the feature pyramid fusion, while a direct concatenation achieves a relative increase ($\uparrow$0.27\%),  the cross-attention mechanism provides a more significant learning opportunity  ($\uparrow$0.70\%).

\begin{table*}[t!]
\scriptsize
\begin{minipage}[t]{0.31\textwidth}
\scriptsize
\vspace{-1.4cm}
\begin{center}
\setlength{\tabcolsep}{1.5mm}
{
\begin{tabular}{|c|ccc|}
\hline
\multicolumn{1}{|c|}{\multirow {2}{*}{\bf \makecell{Centricity\\ Weight $\lambda_{3}$}}} &\multicolumn{3}{c|}{\bf Avg. mAP@task}\\
\cline{2-4}
\multicolumn{1}{|c|}{}&\multicolumn{1}{c}{Verb} &Noun &Action 
\\ 
\hline\hline
0.5 &21.18 &22.82 &18.04
 \\
1   &21.19 &22.02 &18.11
\\
1.5 &21.23 &\textbf{23.24} &18.24
\\
1.6 &\textbf{21.71} &22.89 &18.16
\\
1.7 &21.10 &23.08 &\textbf{18.50}
\\
1.8 &21.16 &22.77 &18.13
\\
2  &21.27 &22.90 &18.22
\\
\hline
\end{tabular}}
\end{center}
\caption{Ablation on varying centricity weight  $\lambda_{3}$.}
\label{Ablations Tab: centricity weight}
\end{minipage}
\hspace{0.1cm}
\begin{minipage}[t]{0.67\textwidth}
\begin{center}
\begin{tabular}{|l|ccc|}
\hline
\multicolumn{1}{|c}{\multirow {2}{*}{\bf Fusion Strategies}} &\multicolumn{3}{|c|}{\bf Avg. mAP@task}\\
\cline{2-4}
\multicolumn{1}{|c}{} &\multicolumn{1}{|c}{Verb} & Noun & Action \\
\hline\hline
Visual-only &20.71 &20.53 &17.18\\
Audio-only &7.94 &6.41 &4.57\\
\hline
Proposals fusion (Figure \ref{fig:diffusion} (a)) &20.36 &20.38 &16.65\\
Classif. scores fusion (multiplication) (Fig \ref{fig:diffusion}(b)) &20.93 &21.82 &17.37\\
Classif. scores fusion (addition) (Fig \ref{fig:diffusion}(b)) &20.18 &\textbf{21.89} &17.69\\
Feature pyramid fusion (concatenation) (Fig \ref{fig:diffusion}(c))&\textbf{20.95}  &21.52  &17.45 \\
Feature pyramid fusion (cross-attention) (Fig \ref{fig:diffusion}(c)) &20.93 &21.84 &\textbf{17.88}\\
\hline
\end{tabular}
\end{center}
\caption{Comparing different strategies to fuse audio \& visual modalities.}
\label{tab:diff fusion}
\end{minipage}
\end{table*}

\begin{figure*}[t!]
\begin{tabular}{cccc}
\bmvaHangBox{\includegraphics[width=0.25\linewidth]{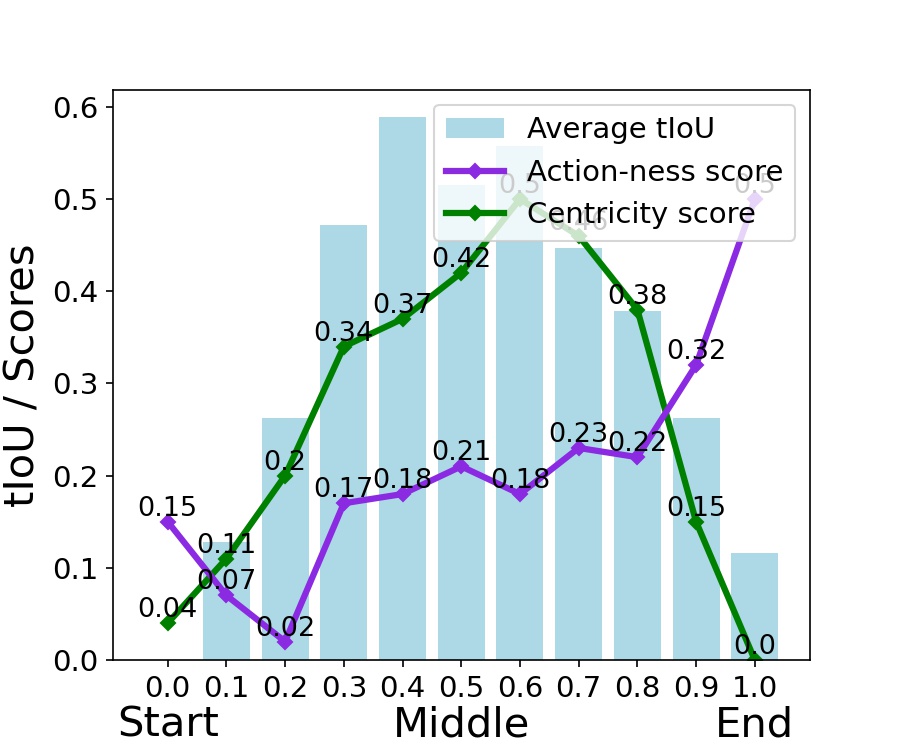}}&\hspace{-5mm}
\bmvaHangBox{\includegraphics[width=0.25\linewidth]{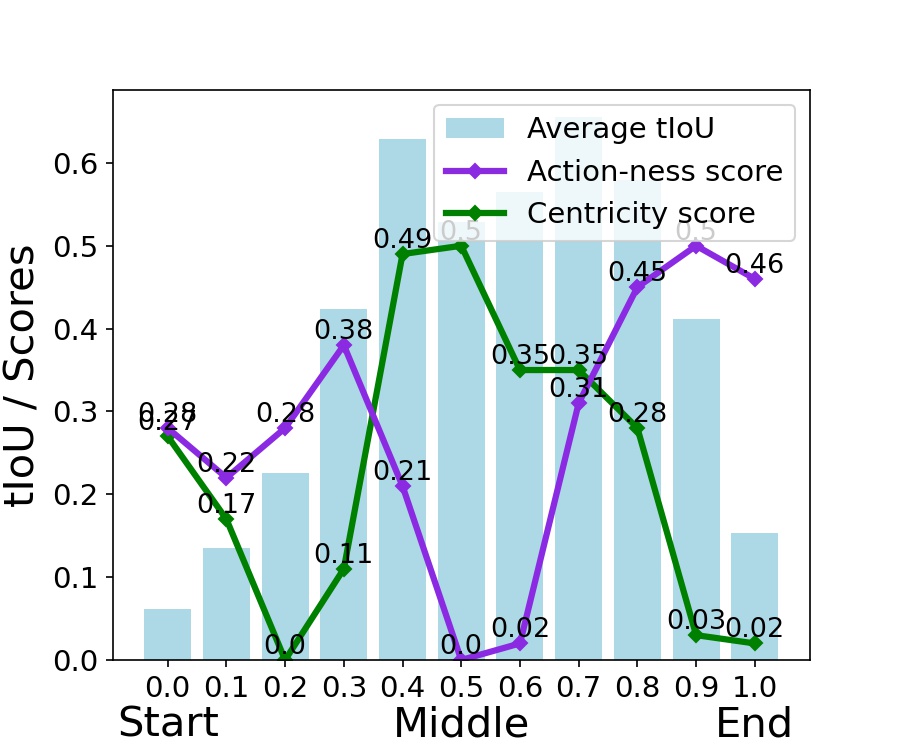}}&\hspace{-5mm}
\bmvaHangBox{\includegraphics[width=0.25\linewidth]{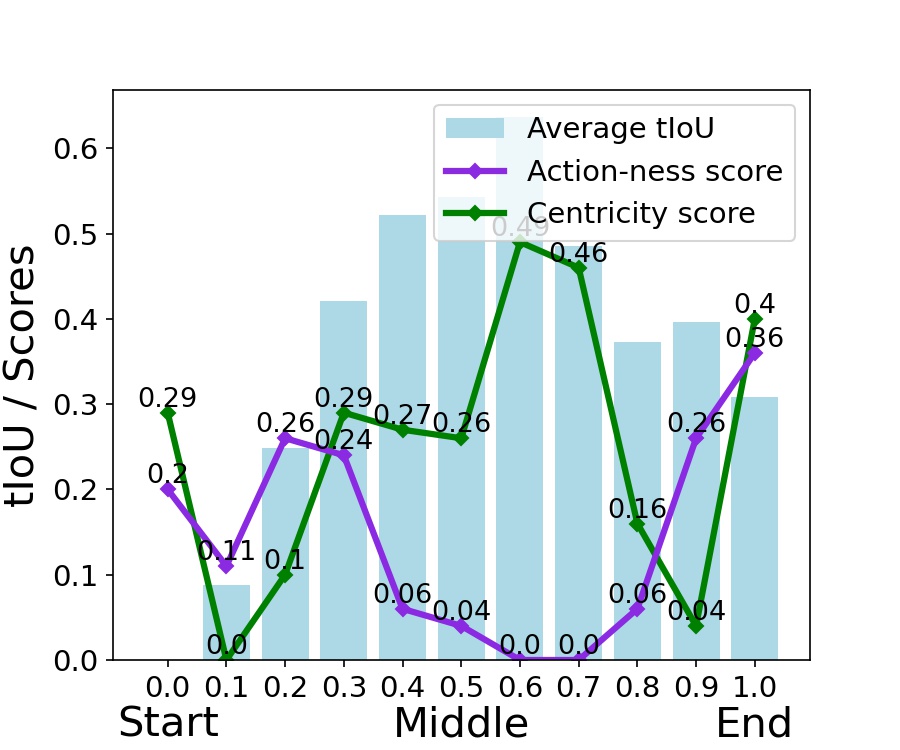}}&\hspace{-5mm}
\bmvaHangBox{\includegraphics[width=0.25\linewidth]{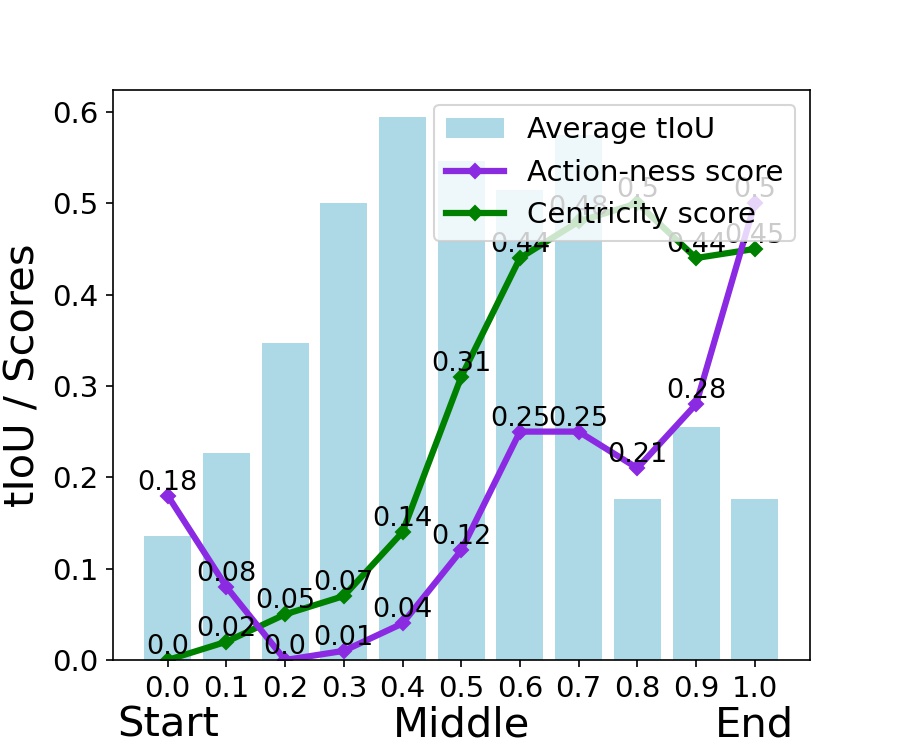}}\\
(a)&(b)&(c)&(d)
\end{tabular}
\vspace*{-3pt}
\caption{Visualization examples of tIoU values, and centricity and action-ness scores -- The x-axis represents the re-scaling of the temporal dimension of an action segment to the range of $[0-1]$. The bars represent the average tIoU between the proposals generated by corresponding timesteps and their ground truth. The green and purple lines are the centricity and action-ness scores, respectively.}
\label{Exmples_tiou_scores}
\end{figure*}

\noindent {\bf Centricity vs. action-ness --} Centricity establishes how close the current timestep is from the action centre, while action-ness \cite{chang2022augmented} represents the probability of the action occurring. Figure \ref{Exmples_tiou_scores} displays individual instances of actions with centricity scores, action-ness scores and the average tIoUs between the proposals generated by corresponding timesteps and their ground truth. 
In the middle of the action, timesteps tend to exhibit peak values of tIoUs that gradually decrease on both sides, and a similar trend is also observed in the centricity scores. This suggests that timesteps associated with higher centricity scores are inclined to generate proposals with more precise boundaries.
In contrast, the action-ness scores (purple line) tend to drop in the middle of the action.

\section{Conclusion}
We introduced an audio-visual fusion approach and a novel centricity head for one-stage anchor-free action detectors. 
Our method achieves state-of-the-art results on the large-scale egocentric EPIC-Kitchens-100 action detection benchmark where audio and video streams are available. Detailed ablations demonstrated the benefits of fusing audio and visual modalities and emphasized the importance of centricity scores. 

Many questions about the use of multi-modalities in temporal action detection remain unexplored, such as the discrepancies in the training data and the temporal misalignment between different modalities.
An extension of our work would involve joint learning from visual, audio, and language modalities to enhance action detection performance, with specific focus on mitigating the misalignment among these three modalities and developing novel fusion techniques to provide discriminative representations.

\bibliography{egbib}

\begin{thebibliography}{50}
\providecommand{\natexlab}[1]{#1}
\providecommand{\url}[1]{\texttt{#1}}
\expandafter\ifx\csname urlstyle\endcsname\relax
  \providecommand{\doi}[1]{doi: #1}\else
  \providecommand{\doi}{doi: \begingroup \urlstyle{rm}\Url}\fi

\bibitem[Afouras et~al.(2020)Afouras, Owens, Chung, and
  Zisserman]{afouras2020self}
Triantafyllos Afouras, Andrew Owens, Joon~Son Chung, and Andrew Zisserman.
\newblock Self-supervised learning of audio-visual objects from video.
\newblock In \emph{Proceedings of the European Conference on Computer Vision
  (ECCV)}, pages 208--224, 2020.

\bibitem[Arandjelovic and Zisserman(2017)]{arandjelovic2017look}
Relja Arandjelovic and Andrew Zisserman.
\newblock Look, listen and learn.
\newblock In \emph{Proceedings of the IEEE/CVF International Conference on
  Computer Vision (ICCV)}, pages 609--617, 2017.

\bibitem[Aytar et~al.(2016)Aytar, Vondrick, and Torralba]{aytar2016soundnet}
Yusuf Aytar, Carl Vondrick, and Antonio Torralba.
\newblock Sound{N}et: Learning sound representations from unlabeled video.
\newblock In \emph{Advances in Neural Information Processing Systems
  (NeurIPS)}, pages 892--900, 2016.

\bibitem[Bodla et~al.(2017)Bodla, Singh, Chellappa, and Davis]{softnms}
Navaneeth Bodla, Bharat Singh, Rama Chellappa, and Larry~S. Davis.
\newblock Soft{-}{NMS} {-} improving object detection with one line of code.
\newblock In \emph{Proceedings of the IEEE/CVF International Conference on
  Computer Vision (ICCV)}, pages 5562--5570, 2017.

\bibitem[Carion et~al.(2020)Carion, Massa, Synnaeve, Usunier, Kirillov, and
  Zagoruyko]{DETR}
Nicolas Carion, Francisco Massa, Gabriel Synnaeve, Nicolas Usunier, Alexander
  Kirillov, and Sergey Zagoruyko.
\newblock End-to-end object detection with transformers.
\newblock In \emph{Proceedings of the European Conference on Computer Vision
  (ECCV)}, pages 213--229, 2020.

\bibitem[Carreira and Zisserman(2017)]{I3D}
Joao Carreira and Andrew Zisserman.
\newblock Quo vadis, action recognition? a new model and the {K}inetics
  dataset.
\newblock In \emph{Proceedings of the IEEE/CVF Conference on Computer Vision
  and Pattern Recognition (CVPR)}, pages 6299--6308, 2017.

\bibitem[Chang et~al.(2022)Chang, Wang, Wang, Li, and Shou]{chang2022augmented}
Shuning Chang, Pichao Wang, Fan Wang, Hao Li, and Zheng Shou.
\newblock Augmented transformer with adaptive graph for temporal action
  proposal generation.
\newblock In \emph{Proceedings of the 3rd International Workshop on
  Human-Centric Multimedia Analysis}, pages 41--50, 2022.

\bibitem[Cheng and Bertasius(2022)]{cheng2022tallformer}
Feng Cheng and Gedas Bertasius.
\newblock Tall{F}ormer: Temporal action localization with a long-memory
  transformer.
\newblock In \emph{Proceedings of the European Conference on Computer Vision
  (ECCV)}, pages 503--521, 2022.

\bibitem[Damen et~al.(2022)Damen, Doughty, Farinella, Furnari, Kazakos, Ma,
  Moltisanti, Munro, Perrett, Price, et~al.]{EPIC100}
Dima Damen, Hazel Doughty, Giovanni~Maria Farinella, Antonino Furnari,
  Evangelos Kazakos, Jian Ma, Davide Moltisanti, Jonathan Munro, Toby Perrett,
  Will Price, et~al.
\newblock Rescaling egocentric vision.
\newblock \emph{International Journal of Computer Vision (IJCV)}, pages 33--55,
  2022.

\bibitem[Feichtenhofer et~al.(2019)Feichtenhofer, Fan, Malik, and He]{slowfast}
Christoph Feichtenhofer, Haoqi Fan, Jitendra Malik, and Kaiming He.
\newblock Slow{F}ast networks for video recognition.
\newblock In \emph{Proceedings of the IEEE/CVF International Conference on
  Computer Vision (ICCV)}, pages 6202--6211, 2019.

\bibitem[Gao et~al.(2017)Gao, Yang, Chen, Sun, and Nevatia]{Turntap}
Jiyang Gao, Zhenheng Yang, Kan Chen, Chen Sun, and Ram Nevatia.
\newblock {TURN TAP}: Temporal unit regression network for temporal action
  proposals.
\newblock In \emph{Proceedings of the IEEE/CVF International Conference on
  Computer Vision (ICCV)}, pages 3628--3636, 2017.

\bibitem[Grauman et~al.(2022)Grauman, Westbury, Byrne, Chavis, Furnari,
  Girdhar, Hamburger, Jiang, Liu, Liu, et~al.]{Ego4d}
Kristen Grauman, Andrew Westbury, Eugene Byrne, Zachary Chavis, Antonino
  Furnari, Rohit Girdhar, Jackson Hamburger, Hao Jiang, Miao Liu, Xingyu Liu,
  et~al.
\newblock Ego4{D}: Around the world in 3,000 hours of egocentric video.
\newblock In \emph{Proceedings of the IEEE/CVF Conference on Computer Vision
  and Pattern Recognition (CVPR)}, pages 18995--19012, 2022.

\bibitem[Hu et~al.(2019)Hu, Nie, and Li]{hu2019deep}
Di~Hu, Feiping Nie, and Xuelong Li.
\newblock Deep multimodal clustering for unsupervised audiovisual learning.
\newblock In \emph{Proceedings of the IEEE/CVF Conference on Computer Vision
  and Pattern Recognition (CVPR)}, pages 9248--9257, 2019.

\bibitem[Huang et~al.(2022)Huang, Zhang, Pan, Qing, Tang, Liu, and
  Ang~Jr]{TADA}
Ziyuan Huang, Shiwei Zhang, Liang Pan, Zhiwu Qing, Mingqian Tang, Ziwei Liu,
  and Marcelo~H Ang~Jr.
\newblock T{A}da! temporally-adaptive convolutions for video understanding.
\newblock In \emph{Proceedings of International Conference on Learning
  Representations (ICLR)}, pages 1--23, 2022.

\bibitem[Huh et~al.(2023)Huh, Chalk, Kazakos, Damen, and
  Zisserman]{EPICSOUNDS2023}
Jaesung Huh, Jacob Chalk, Evangelos Kazakos, Dima Damen, and Andrew Zisserman.
\newblock {EPIC-SOUNDS}: {A} {L}arge-{S}cale {D}ataset of {A}ctions that
  {S}ound.
\newblock In \emph{IEEE International Conference on Acoustics, Speech and
  Signal Processing (ICASSP)}, 2023.

\bibitem[Kazakos et~al.(2019)Kazakos, Nagrani, Zisserman, and
  Damen]{kazakos2019epic}
Evangelos Kazakos, Arsha Nagrani, Andrew Zisserman, and Dima Damen.
\newblock {EPIC}-{F}usion: Audio-visual temporal binding for egocentric action
  recognition.
\newblock In \emph{Proceedings of the IEEE/CVF International Conference on
  Computer Vision (ICCV)}, pages 5492--5501, 2019.

\bibitem[Kazakos et~al.(2021{\natexlab{a}})Kazakos, Huh, Nagrani, Zisserman,
  and Damen]{kazakos2021MTCN}
Evangelos Kazakos, Jaesung Huh, Arsha Nagrani, Andrew Zisserman, and Dima
  Damen.
\newblock With a little help from my temporal context: Multimodal egocentric
  action recognition.
\newblock In \emph{Proceedings of the British Machine Vision Conference
  (BMVC)}, 2021{\natexlab{a}}.

\bibitem[Kazakos et~al.(2021{\natexlab{b}})Kazakos, Nagrani, Zisserman, and
  Damen]{kazakos2021slow}
Evangelos Kazakos, Arsha Nagrani, Andrew Zisserman, and Dima Damen.
\newblock Slow-{F}ast auditory streams for audio recognition.
\newblock In \emph{IEEE International Conference on Acoustics, Speech and
  Signal Processing (ICASSP)}, pages 855--859. IEEE, 2021{\natexlab{b}}.

\bibitem[Korbar et~al.(2018)Korbar, Tran, and Torresani]{korbar2018cooperative}
Bruno Korbar, Du~Tran, and Lorenzo Torresani.
\newblock Cooperative learning of audio and video models from self-supervised
  synchronization.
\newblock \emph{Advances in Neural Information Processing Systems (NeurIPS)},
  pages 7774--7785, 2018.

\bibitem[Lee et~al.(2021)Lee, Jain, Park, and Yun]{lee2021cross}
Jun-Tae Lee, Mihir Jain, Hyoungwoo Park, and Sungrack Yun.
\newblock Cross-attentional audio-visual fusion for weakly-supervised action
  localization.
\newblock In \emph{Proceedings of International Conference on Learning
  Representations (ICLR)}, 2021.

\bibitem[Li et~al.(2016)Li, Lan, Xing, Zeng, Yuan, and Liu]{Gaussian2016}
Yanghao Li, Cuiling Lan, Junliang Xing, Wenjun Zeng, Chunfeng Yuan, and Jiaying
  Liu.
\newblock Online human action detection using joint classification-regression
  recurrent neural networks.
\newblock In \emph{Proceedings of the European Conference on Computer Vision
  (ECCV)}, pages 203--220, 2016.

\bibitem[Lin et~al.(2020)Lin, Li, Wang, Tai, Luo, Cui, Wang, Li, Huang, and
  Ji]{DBG}
Chuming Lin, Jian Li, Yabiao Wang, Ying Tai, Donghao Luo, Zhipeng Cui, Chengjie
  Wang, Jilin Li, Feiyue Huang, and Rongrong Ji.
\newblock Fast learning of temporal action proposal via dense boundary
  generator.
\newblock In \emph{Proceedings of the AAAI Conference on Artificial
  Intelligence (AAAI)}, pages 11499--11506, 2020.

\bibitem[Lin et~al.(2021)Lin, Xu, Luo, Wang, Tai, Wang, Li, Huang, and
  Fu]{AFSD}
Chuming Lin, Chengming Xu, Donghao Luo, Yabiao Wang, Ying Tai, Chengjie Wang,
  Jilin Li, Feiyue Huang, and Yanwei Fu.
\newblock Learning salient boundary feature for anchor-free temporal action
  localization.
\newblock In \emph{Proceedings of the IEEE/CVF Conference on Computer Vision
  and Pattern Recognition (CVPR)}, pages 3320--3329, 2021.

\bibitem[Lin et~al.(2017{\natexlab{a}})Lin, Zhao, and Shou]{SSAD}
Tianwei Lin, Xu~Zhao, and Zheng Shou.
\newblock Single shot temporal action detection.
\newblock In \emph{Proceedings of the 25th ACM International Conference on
  Multimedia}, pages 988--996, 2017{\natexlab{a}}.

\bibitem[Lin et~al.(2018)Lin, Zhao, Su, Wang, and Yang]{BSN}
Tianwei Lin, Xu~Zhao, Haisheng Su, Chongjing Wang, and Ming Yang.
\newblock {BSN}: Boundary sensitive network for temporal action proposal
  generation.
\newblock In \emph{Proceedings of the European Conference on Computer Vision
  (ECCV)}, pages 3--19, 2018.

\bibitem[Lin et~al.(2019)Lin, Liu, Li, Ding, and Wen]{BMN}
Tianwei Lin, Xiao Liu, Xin Li, Errui Ding, and Shilei Wen.
\newblock {BMN}: Boundary-matching network for temporal action proposal
  generation.
\newblock In \emph{Proceedings of the IEEE/CVF International Conference on
  Computer Vision (ICCV)}, pages 3889--3898, 2019.

\bibitem[Lin et~al.(2017{\natexlab{b}})Lin, Goyal, Girshick, He, and
  Doll{\'a}r]{focal}
Tsung-Yi Lin, Priya Goyal, Ross Girshick, Kaiming He, and Piotr Doll{\'a}r.
\newblock Focal loss for dense object detection.
\newblock In \emph{Proceedings of the IEEE/CVF International Conference on
  Computer Vision (ICCV)}, pages 2980--2988, 2017{\natexlab{b}}.

\bibitem[Liu and Wang(2020)]{PBRNet}
Qinying Liu and Zilei Wang.
\newblock Progressive boundary refinement network for temporal action
  detection.
\newblock In \emph{Proceedings of the AAAI Conference on Artificial
  Intelligence (AAAI)}, pages 11612--11619, 2020.

\bibitem[Liu et~al.(2022)Liu, Wang, Hu, Tang, Zhang, Bai, and Bai]{TadTR}
Xiaolong Liu, Qimeng Wang, Yao Hu, Xu~Tang, Shiwei Zhang, Song Bai, and Xiang
  Bai.
\newblock End-to-end temporal action detection with transformer.
\newblock \emph{IEEE Transactions on Image Processing}, pages 5427--5441, 2022.

\bibitem[Long et~al.(2019)Long, Yao, Qiu, Tian, Luo, and Mei]{GTAN}
Fuchen Long, Ting Yao, Zhaofan Qiu, Xinmei Tian, Jiebo Luo, and Tao Mei.
\newblock Gaussian temporal awareness networks for action localization.
\newblock In \emph{Proceedings of the IEEE/CVF Conference on Computer Vision
  and Pattern Recognition (CVPR)}, pages 344--353, 2019.

\bibitem[Mo and Tian(2022)]{vid_parsing_2022}
Shentong Mo and Yapeng Tian.
\newblock Multi-modal grouping network for weakly-supervised audio-visual video
  parsing.
\newblock In \emph{Advances in Neural Information Processing Systems
  (NeurIPS)}, pages 34722--34733, 2022.

\bibitem[Nagrani et~al.(2020)Nagrani, Sun, Ross, Sukthankar, Schmid, and
  Zisserman]{Nagrani20c}
Arsha Nagrani, Chen Sun, David Ross, Rahul Sukthankar, Cordelia Schmid, and
  Andrew Zisserman.
\newblock Speech2{A}ction: Cross-modal supervision for action recognition.
\newblock In \emph{Proceedings of the IEEE/CVF Conference on Computer Vision
  and Pattern Recognition (CVPR)}, pages 10317--10326, 2020.

\bibitem[Nawhal and Mori(2021)]{AGT}
Megha Nawhal and Greg Mori.
\newblock Activity graph transformer for temporal action localization.
\newblock \emph{arXiv preprint arXiv:2101.08540}, 2021.

\bibitem[Owens and Efros(2018)]{owens2018audio}
Andrew Owens and Alexei~A Efros.
\newblock Audio-visual scene analysis with self-supervised multisensory
  features.
\newblock In \emph{Proceedings of the European Conference on Computer Vision
  (ECCV)}, pages 631--648, 2018.

\bibitem[Ramazanova et~al.(2022)Ramazanova, Escorcia, Heilbron, Zhao, and
  Ghanem]{OWL}
Merey Ramazanova, Victor Escorcia, Fabian~Caba Heilbron, Chen Zhao, and Bernard
  Ghanem.
\newblock O{WL} ({O}bserve, {W}atch, {L}isten): Localizing actions in
  egocentric video via audiovisual temporal context.
\newblock \emph{arXiv preprint arXiv:2202.04947}, 2022.

\bibitem[Rao et~al.(2022)Rao, Khalil, Li, Dai, and Lu]{rao2022dual}
Varshanth Rao, Md~Ibrahim Khalil, Haoda Li, Peng Dai, and Juwei Lu.
\newblock Dual perspective network for audio-visual event localization.
\newblock In \emph{Proceedings of the European Conference on Computer Vision
  (ECCV)}, pages 689--704, 2022.

\bibitem[Rezatofighi et~al.(2019)Rezatofighi, Tsoi, Gwak, Sadeghian, Reid, and
  Savarese]{iouloss}
Hamid Rezatofighi, Nathan Tsoi, JunYoung Gwak, Amir Sadeghian, Ian Reid, and
  Silvio Savarese.
\newblock Generalized intersection over union: A metric and a loss for bounding
  box regression.
\newblock In \emph{Proceedings of the IEEE/CVF Conference on Computer Vision
  and Pattern Recognition (CVPR)}, pages 658--666, 2019.

\bibitem[Shi et~al.(2022)Shi, Zhong, Cao, Zhang, Ma, Li, and Tao]{shi2022react}
Dingfeng Shi, Yujie Zhong, Qiong Cao, Jing Zhang, Lin Ma, Jia Li, and Dacheng
  Tao.
\newblock Re{A}ct: Temporal action detection with relational queries.
\newblock In \emph{Proceedings of the European Conference on Computer Vision
  (ECCV)}, pages 105--121, 2022.

\bibitem[Shi et~al.(2023)Shi, Zhong, Cao, Ma, Li, and Tao]{Tridet}
Dingfeng Shi, Yujie Zhong, Qiong Cao, Lin Ma, Jia Li, and Dacheng Tao.
\newblock Tri{D}et: Temporal action detection with relative boundary modeling.
\newblock \emph{Proceedings of the IEEE/CVF Conference on Computer Vision and
  Pattern Recognition (CVPR)}, pages 18857--18866, 2023.

\bibitem[Su et~al.(2021)Su, Gan, Wu, Yan, and Qiao]{BSN++}
Haisheng Su, Weihao Gan, Wei Wu, Junjie Yan, and Yu~Qiao.
\newblock {BSN++}: Complementary boundary regressor with scale-balanced
  relation modeling for temporal action proposal generation.
\newblock In \emph{Proceedings of the AAAI Conference on Artificial
  Intelligence (AAAI)}, pages 2602--2610, 2021.

\bibitem[Tian et~al.(2018)Tian, Shi, Li, Duan, and Xu]{AVE}
Yapeng Tian, Jing Shi, Bochen Li, Zhiyao Duan, and Chenliang Xu.
\newblock Audio-visual event localization in unconstrained videos.
\newblock In \emph{Proceedings of the European Conference on Computer Vision
  (ECCV)}, pages 247--263, 2018.

\bibitem[Wang et~al.(2022)Wang, Mirmehdi, Damen, and Perrett]{wang2022refining}
Hanyuan Wang, Majid Mirmehdi, Dima Damen, and Toby Perrett.
\newblock Refining action boundaries for one-stage detection.
\newblock In \emph{18th IEEE International Conference on Advanced Video and
  Signal Based Surveillance (AVSS)}, pages 1--8. IEEE, 2022.

\bibitem[Wang et~al.(2023)Wang, Singh, and Torresani]{wang2023ego}
Huiyu Wang, Mitesh~Kumar Singh, and Lorenzo Torresani.
\newblock Ego-{O}nly: Egocentric action detection without exocentric
  pretraining.
\newblock \emph{arXiv preprint arXiv:2301.01380}, 2023.

\bibitem[Wang et~al.(2016)Wang, Xiong, Wang, Qiao, Lin, Tang, and
  Van~Gool]{TSN}
Limin Wang, Yuanjun Xiong, Zhe Wang, Yu~Qiao, Dahua Lin, Xiaoou Tang, and Luc
  Van~Gool.
\newblock Temporal segment networks: Towards good practices for deep action
  recognition.
\newblock In \emph{Proceedings of the European Conference on Computer Vision
  (ECCV)}, pages 20--36, 2016.

\bibitem[Wu and Yang(2021)]{wu2021exploring}
Yu~Wu and Yi~Yang.
\newblock Exploring heterogeneous clues for weakly-supervised audio-visual
  video parsing.
\newblock In \emph{Proceedings of the IEEE/CVF Conference on Computer Vision
  and Pattern Recognition (CVPR)}, pages 1326--1335, 2021.

\bibitem[Xia and Zhao(2022)]{xia2022cross}
Yan Xia and Zhou Zhao.
\newblock Cross-modal background suppression for audio-visual event
  localization.
\newblock In \emph{Proceedings of the IEEE/CVF Conference on Computer Vision
  and Pattern Recognition (CVPR)}, pages 19989--19998, 2022.

\bibitem[Xu et~al.(2017)Xu, Das, and Saenko]{RC3D}
Huijuan Xu, Abir Das, and Kate Saenko.
\newblock R-{C3D}: Region convolutional 3d network for temporal activity
  detection.
\newblock In \emph{Proceedings of the IEEE/CVF International Conference on
  Computer Vision (ICCV)}, pages 5783--5792, 2017.

\bibitem[Yang et~al.(2020)Yang, Peng, Zhang, Fu, and Han]{A2Net}
Le~Yang, Houwen Peng, Dingwen Zhang, Jianlong Fu, and Junwei Han.
\newblock Revisiting anchor mechanisms for temporal action localization.
\newblock \emph{IEEE Transactions on Image Processing}, pages 8535--8548, 2020.

\bibitem[Zhang et~al.(2022)Zhang, Wu, and Li]{zhang2022actionformer}
Chen-Lin Zhang, Jianxin Wu, and Yin Li.
\newblock Action{F}ormer: Localizing moments of actions with transformers.
\newblock In \emph{Proceedings of the European Conference on Computer Vision
  (ECCV)}, pages 492--510, 2022.

\bibitem[Zhao et~al.(2017)Zhao, Xiong, Wang, Wu, Tang, and Lin]{SSN}
Yue Zhao, Yuanjun Xiong, Limin Wang, Zhirong Wu, Xiaoou Tang, and Dahua Lin.
\newblock Temporal action detection with structured segment networks.
\newblock In \emph{Proceedings of the IEEE/CVF International Conference on
  Computer Vision (ICCV)}, pages 2914--2923, 2017.

\end{thebibliography}
\end{document}